# Colorectal cancer diagnosis from histology images: A comparative study

Junaid Malik, Serkan Kiranyaz, Suchitra Kunhoth, Turker Ince, Somaya Al-Maadeed, Ridha Hamila, Moncef Gabbouj

*Abstract*—Computer-aided diagnosis (CAD) based on histopathological imaging has progressed rapidly in recent years with the rise of machine learning based methodologies. Traditional approaches consist of training a classification model using features extracted from the images, based on textures or morphological properties. Recently, deep-learning based methods have been applied directly to the raw (unprocessed) data. However, their usability is impacted by the paucity of annotated data in the biomedical sector. In order to leverage the learning capabilities of deep Convolutional Neural Nets (CNNs) within the confines of limited labelled data, in this study we shall investigate the transfer learning approaches that aim to apply the knowledge gained from solving a source (e.g., non-medical) problem, to learn better predictive models for the target (e.g., biomedical) task. As an alternative, we shall further propose a new adaptive and compact CNN based architecture that can be trained from scratch even on scarce and low-resolution data. Moreover, we conduct quantitative comparative evaluations among the traditional methods, transfer learning-based methods and the proposed adaptive approach for the particular task of cancer detection and identification from scarce and low-resolution histology images. Over the largest benchmark dataset formed for this purpose, the proposed adaptive approach achieved a higher cancer detection accuracy with a significant gap, whereas the deep CNNs with transfer learning achieved a superior cancer identification.

## I. INTRODUCTION

According to recent statistics, cancer is attributed to approximately 8.8 million deaths worldwide, making it the second most deadliest disease and the primary cause of nearly one in six deaths globally [1]. Ranked based on incidence rates, colorectal cancer (CRC) is the 3$^{rd}$ most common form of cancer [2] preceding only by lung and breast cancers, respectively . For an effective treatment, timely detection and classification of the cancerous cells is imperative [3] as, according to the American Cancer Society, 56% of the patients with colorectal cancer are diagnosed at regional or distant stage, whereby the cancer has started to spread out from the primary tumor to other parts of the body [4]. Rapid technological advancements in the field of image processing and machine learning have led to the introduction of numerous cost-effective and fast computer-aided diagnostic methodologies. Traditional methods generally aim to perform pattern recognition-based systems for fast and automated cancer diagnosis. This involves extracting a fixed set of hand-crafted features from the histology images based on, e.g., texture and morphological properties, and training a classifier over these features to classify/detect cancerous cells. Recently, the field has seen the advent of deep artificial neural networks, which combine the feature extraction and classification within a unified learning body.

Conventional (deep) Convolutional Neural Networks (CNNs) are feed-forward artificial neural networks inspired by the mammalian visual cortex. Deep CNNs, compromising of numerous hidden layers, have become the *de-facto* standard for many visual recognition applications (e.g., object recognition, segmentation, tracking, etc.) as they achieved the state-of-the-art performance [5], [6] with a significant performance gap. However, in order to train these deep CNNs from scratch, a massive size dataset in the scale of "Big-Data" is usually required with the ground-truth labels (annotations). This may not be feasible especially for biomedical datasets where the labelled data is scarce and usually limited due to the privacy issues. Moreover, training is also expensive in terms of time and computational resources. Finally for biomedical image expert annotations are very costly and laborious [7]. These factors paralyze the application of deep CNNs in many practical problems.

Transfer learning has gained popularity among researchers as a viable alternative to training deep CNNs from scratch. It commonly involves re-using a CNN model that has been trained on a large but otherwise unrelated dataset. The original architecture of the pre-trained model is retained and is used, either as a feature extractor, or is fine-tuned by resuming training using the available data. Despite the apparent disparity among natural images and biomedical imaging data, [8] showed that deep CNN models trained on large-scale natural image datasets can effectively be fine-tuned to obtain consistent and accurate results for biomedical image analysis.

Another prevalent approach to alleviate the issue of limited datasets is using augmentation and enrichment techniques to

Junaid Malik (hafiz.malik@qu.edu.qa), Serkan Kiranyaz (mkiranyaz@qu.edu.qa), Suchitra Kunhoth (suchithra@qu.edu.qa), Somaya Al-Maadeed (s_alali@qu.edu.qa) and Ridha Hamila (hamila@qu.edu.qa) are with the Department of Engineering, Qatar University, Doha, Qatar.

Turker Ince (turker.ince@ieu.edu.tr) is with Izmir University of Economics.

Moncef Gabbouj (moncef.gabbouj@tut.fi) is with the Laboratory of Signal Processing, Tampere University of Technology, Finland.

refine and virtually enrich the available (train) data. In histopathological analysis, different patch generation and data transformation techniques are often employed. Data enrichment is also applied by incorporating a segmentation stage as a precursor to feature extraction and classification [7], [9]–[11]. This involves removing the background and emphasizing regions of interest by localizing tumor areas. However, the extent up to which such transformations can add meaningful information on top of the available data is limited. Furthermore, most of the preprocessing techniques rely on manual or semi-automated user-dependent procedures, which introduce considerable subjectivity and feasibility issues. Another critical factor in this regard is the limited variability in the dataset; identified by the number of unique patients from which the data was collected. Even in the datasets that contain a relatively high number of images [11]–[13], the number of patients is low. Such a lack of diversity hinders proper validation and it cannot be remedied by any data augmentation or enrichment procedure.

Considering the aforementioned facts, in this study, we investigate and compare the effectiveness of traditional and novel CNN-based methods for the particular task of colorectal cancer screening from scarce and low-resolution histology images. Specifically, we aim to answer the following questions:

- *Which machine learning based approach is best suited for this problem?* To quantify this, results of four CNN-based approaches are compared against five traditional classification methods, each based on a state-of-the-art texture feature.
- *Are the features learnt from large-scale natural image datasets transferrable to cancer diagnosis?* To this end, we use the state-of-the-art InceptionV3 architecture, that has been trained on ImageNet [14] dataset, and fine-tune it for the task of interest.
- *Does training a compact CNN from scratch provide a better alternative to fine-tuning deep pre-trained models?* For this, we propose a systematic approach using a compact and adaptive CNN and compare its results with the fine-tuned models.

We keep all our experiments devoid of any prior manual or semi-automatic image enhancement or segmentation stages. Furthermore, to tackle the issue of variability within the dataset, we gather and employ biopsy samples of 151 patients, equally distributed among each of 4 classes. This has, as of yet, became the largest benchmark dataset with the highest number of patients ever used for this challenging task.

The rest of the paper is organized as follows. Section II provides a brief review of the related work. Section III describes the benchmark dataset used in our study along with the adapted data augmentation approach. Section IV presents the overview of the comparative evaluation performed. The experimental results are further presented in Section V. Finally, Section VI concludes the paper.

## II. PRIOR WORK

Traditional supervised approaches for the task of cancer screening generally involve a feature extraction step followed by training a classifier over these features. For the binary colorectal cancer detection problem, the authors of [15] investigated various combinations of 18 extracted features, including textural features derived from the grey level co-occurrence matrix (GLCM). A support vector machine (SVM) based classifier was then trained which achieved a top detection precision of 96.67% on a dataset compromising of 60 images equally distributed between the two classes. A similar investigative study was conducted in [16] by evaluating a variety of textural features along with different classifiers. The feature extraction step was supplemented by a dimensionality reduction operation based on Principal Component Analysis (PCA). Classifiers were trained for a three-class classification problem using features extracted from a dataset consisting of 29 multispectral images; each containing 16 channels corresponding to different wavelengths. Multiscale local binary patterns (LBP) followed by an SVM-based classifier provided the best classification accuracy. In [17], a large set of features are calculated over a dataset of 48 breast histology images using nuclear architectural characteristics in addition to the commonly used texture-based features. For the cancer detection task, an SVM-based classifier trained on Gabor filter features exhibited top results with 95.8% accuracy. To ensure the extraction of highly discriminative features, the feature extraction step is often preceded by a segmentation operation to emphasize regions of interest. In [10], a derivative of the active contour-based segmentation (snake) method based on progressive subsampling of the image is introduced for the segmentation of multispectral images. Afterwards, Haralick-based features are extracted, and a probabilistic neural network is trained for the four-class cancer classification task. Features from 18 images were used for training the network and a further 45 images were used for testing. Similarly, in [11], an active-contour based preprocessing step was applied to identify regions of interest, from which various textural features are extracted. Three different classifiers were trained using these features and evaluated across a dataset of 480 16-channel multispectral images obtained from biopsy samples of 30 patients. For the task of melanoma skin cancer detection, the study in [18] employs a multilayer perceptron as the classifier fed by 12 highest scoring GLCM features, ranked with respect to Fisher score. Dataset consisted of 102 equally distributed dermascopy images from which 75% were used for training and the rest for testing. The authors report a detection accuracy of 92% for the test set. On a dataset of 200 dermascopy images, the authors of [9] applied an automatic segmentation based pre-processing step followed by a two-level classifier where the images are first classified into normal and abnormal classes, and the abnormal images are further identified as being either atypical or melanoma. Both color and texture features were used to achieve 93.2% classification accuracy of. The study in [19] found that the shape-based features computed on binary segmented cells have high discriminative abilities for carcinoma cell detection. In [12], a combination of conventional textural and morphological features with some new proposed geometric features achieves a 98% accuracy for colorectal cancer detection task over 174 images.

Deep CNNs provide state-of-the-art results for a variety of computer vision tasks such as object detection and recognition [5], [20], [21]. However, a proper training requires a massive number of images, which critically impairs their usability for biomedical imaging problems where the labeled data is often scarce. For the specific case of histopathological image analysis, patch generation and data augmentation techniques are the most common approaches employed to artificially increase the number of labeled data samples used for training. Patch generation involves extracting smaller sized patches from the original image. Each of the patch is considered to have the same label as the original image and is treated as a unique data point. This is often followed by a data augmentation step, which involves applying geometrical transformations such as rotation and flipping to further increase enrich the training data. In [22], 70000 unique patches are generated from 250 original training images, which are then used to train a CNN for 4-class breast cancer classification. Training the CNN for classification and training a classifier using CNN features both achieved around 80% cancer diagnosis accuracy, with the latter producing slightly better results for both binary detection and 4-class classification. Similarly, in [13], a CNN based on the classical architecture of AlexNet [5] is trained from scratch using the BreaKHis [23] dataset. Up to 1000 patches were generated from each of the 7909 images and a maximum image-wise accuracy of 90% was reported. On the same dataset, the authors of [24] cropped each image to obtain a square patch and applied affine transformations to obtain a larger pool of training samples. This was then used to train a custom CNN architecture consisting of three convolution blocks, 3 fully-connected layers followed by a softmax classification layer. This approach outperformed handcrafted features-based methods and achieved an average recognition rate of around 83.25%.

Unlike breast cancer diagnosis, colorectal cancer screening problem lacks a comprehensive dataset like [23]. In [7], multispectral 16-channel images from only 30 patients were obtained and segmented using the snakes method. The segmented images were then used to train a CNN compromising of 2 convolutional layers (each followed by max-pooling) and one fully-connected layer. Although an accuracy of 99% is reported, there is notable subjectivity introduced because of the preprocessing segmentation step and the limited variability of the test set, which contained images from only 9 patients.

As alluded to in the introduction, training a deep CNN from scratch over a limited amount of labelled data is not feasible and the extent to which patch generation and data augmentation alleviate this problem remains limited. As a workaround, transfer learning-based methodologies are introduced to compensate for the scarcity of data. In the realm of biomedical image analysis, [8] explored the application of transfer learning for thoraco-abdominal lymph node detection and interstitial lung disease classification. Models for popular CNN architectures of AlexNet [5] and GoogLeNet [25], that were pre-trained on the large-scale ImageNet [14] dataset, were used for this purpose. These pre-trained models were tested both by using them as feature extractors and by fine-tuning through training only the last fully connected layer from scratch and resuming training for the bottom convolutional layers using a slightly lower learning rate. Transfer learning-based approaches were found to be visually and quantitatively more stable and superior [8]. For the specific task of cancer classification from histology images, several methods based on transfer learning have been proposed recently. Binary classification of histology images for detecting breast cancer was performed in [26]. The authors propose to apply a two-step transfer learning procedure on the dataset of [23]. Following patch generation and data augmentation, a VGG16 model that has been pre-trained on ImageNet dataset is used for transfer learning. Specifically, the fully connected layers of the model are randomly initialized and freshly trained using the target dataset. Afterwards, the whole network is fine-tuned with a comparatively smaller learning rate over the same training data. This transfer learning-based strategy was shown to outperform the methods employing training from scratch for the same set of images. The same cancer detection task was tackled in [27] on the BCDR [28] dataset, which consists of digitized mammograms belonging to more than 300 patients. Several fine-tuning techniques were investigated on a pre-trained Inception V3 model. In addition, a dynamic learning rate adjustment strategy during fine-tuning is presented, whereby the learning rate for each lower layer is exponentially decreased. On a dataset of 400 high resolution images belonging to 4 different classes of breast cancer, [29] applied a stain normalization on the patches before using them to fine-tune pre-trained InceptionV3 and ResNet50 models. The top fully connected layers of the networks were replaced by a combination of average pooling, a fully connected layer followed by a *softmax* classification layer. Fine-tuning was performed over the whole model with a low learning rate using stochastic gradient descent optimizer. An image-wise classification accuracy of more than 90% was reported for both fine-tuned models, with ResNet50 providing slightly better results as compared to Inception V3. On the same dataset, the authors of [30] fine-tuned an Inception-Resnet v2 model, which is an extension of Inception V3 inspired by the residual connections introduced in ResNet50 [31]. Training proceeded by first freezing learning for all convolutional layers and then resuming training for a selected few of them. It achieved 76% accuracy on the test set.

### III. QU-AHLI BENCHMARK DATASET

*A. Colorectal Biopsy samples*

For the dataset used in this study, colorectal tissue slides, and the concerned medical details were obtained from the Pathology and Laboratory Medicine lab at Al-Ahli hospital, Qatar. Qatar University's Institutional Review Board (QU-IRB) provided further review and approval for the study. A total of 164 tissue samples collected between the years 2007 and 2016 were provided by the hospital from 151 different patients. The biopsied tissues were stained with Hematoxylin & Eosin stain. Each sample belongs to one of the four classes of colorectal tissue: Normal, Hyperplastic polyp (HP), Tubular Adenoma with low-grade dysplasia (TA_LG) and Carcinoma (CA). Except for the normal class, the consultant histopathologist did specific marking in order to indicate the ROI belonging to the concerned class. Normal class stands for the non-cancerous, non-malignant class without any disease symptoms [32]. Hyperplastic polyp belongs to the nonneoplastic polyp category. Although they are not dangerous, some hyperplastic polyps on the right side of the colon can be the precursors of the colorectal carcinoma. Tubular Adenoma falls under the neoplastic polyp category. The malignancy with the adenomatous polyp may depend on the polyp size, histologic architecture and the severity of epithelial dysplasia. The adenomatous polyps may transform into the malignant carcinoma if not cured at an early stage.

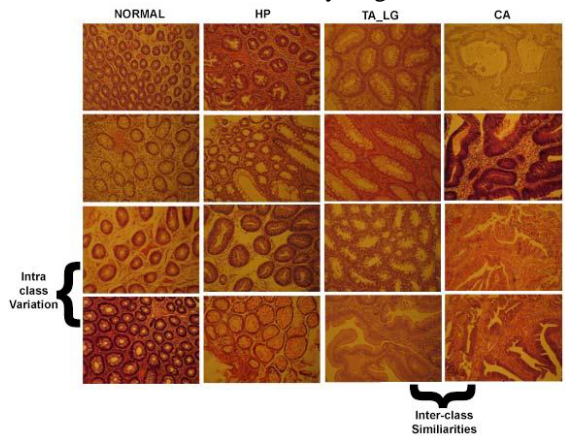

Figure 1 Sample patches from the QU-AHLI Dataset

The biopsy slides were digitized using a Canon PowerShot A650IS digital camera mounted on top of a microscope. The resolution of the obtained images was 640x480 pixels. Furthermore, a halogen illuminated Zeiss microscope was used for the zooming of the specimens. Acquisition was performed with a fixed zoom setting of a 10x on the objective lens. The dataset consists of 200 images, containing 50 images from each of the four classes. Sample images taken from each class are illustrated in Figure 1.

### B. Data Augmentation

As is common practice when dealing with histopathological images, we applied a data augmentation procedure on the existing dataset in order to improve the generalization ability of the classifier with a training over a larger dataset, and mainly to improve the classification accuracy of each image by considering the classification results of all its patches. The 640x480 image was initially divided into four non-overlapping patches of size 300x300; two from the top half and two from the bottom half. Each of the four patches was then subjected to 3 major rotations (90o, 180o and 270o) and a transpose. Hence, as a result of patch extraction and data augmentation, 20 unique patches were extracted from each image. This enabled us to build a database of 4000 images (patches) with 1000 patches per class. Finally, for the CNN-based methods, each patch is down-sampled to 64x64 pixels in order to mimic the low-resolution data. Figure 2 illustrates this process.

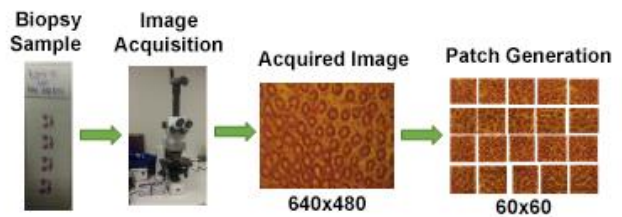

Figure 2 Image acquisition pipeline

### IV. METHODS

In this study, we perform an extensive set of comparative evaluations among three distinct approaches: 5 *state-of-the-art* traditional methods, two recent versions of deep CNNs with transfer learning, and the proposed adaptive and compact CNN trained from scratch over the QU-AHLI dataset. In the following sections, we shall detail each of three approaches.

### A. Traditional Approaches

Traditional machine-learning based methods essentially involve a sequence of two independent steps; feature extraction and classification. We investigate combinations of 5 state-of-the-art feature extraction techniques followed by an SVM classifier with different kernels. Following is a brief description of each of these two steps:

#### 1) Feature Extraction

Local Binary Pattern (LBP) [33] is a local descriptor which encodes the intensity variation between a pixel and its neighbors. Comparison of each pixel with its 8-connected neighbors yields an 8-bit binary codeword, which is then converted to decimal. A histogram is then produced, based on the distribution of these decimal numbers, which is used as a feature vector. Variations of LBP include multiscale LBP (MLBP), rotation invariant LBP (rLBP) and uniform rotation invariant LBP.

Haralick [34] proposed a set of features based on various statisitics that are calculated directly from the GLCM. They are often used as textural descriptors in histopathological image analysis.

Local Phase Quantization [35] is a feature descriptor based on quantized phase of Fourier transform. For each pixel, Fourier transform is computed in the local neighborhood and a label is obtained based on the phase information of the resulting Fourier coefficients. Labels for all image elements are then concatenated to form a histogram, as in the case of LBP, and used as a feature vector. The LPQ codes have been observed to be more robust to blurring. A rotation invariant extension to the original descriptor (rLPQ) as proposed in [36] is deemed to be more effective when dealing with intricate textures such as tissue patterns. Using rLPQ on multispectral images, the authors of [37] reported high performance on tumor cell recognition. Furthermore, the combination of frequency information encoded in LPQ with the spatial information provided by LBP has been proven to be more robust for recognition applications. In this study, we have further investigated the effect of combination of multiple features such as rLPQ and rLBP [38].

*2)   Classification:*
An SVM-based classifier has been found to be most suitable by numerous earlier studies dealing with histology images [39]. In our work, we use an SVM classifier and investigate three different kernels to maximize the classification performace: Radial Basis Functions (RBF), linear and polynomial kernels. Overall, we aim to find the best combination of the texture features and of the SVM kernel together which yields the top classification performance.

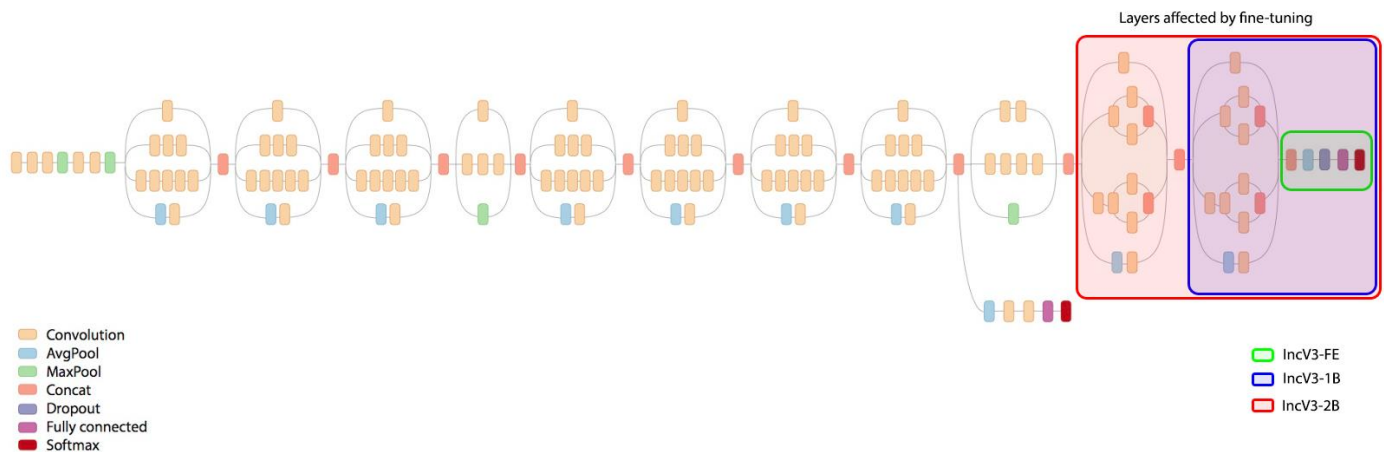

Figure 3 Inception v3 [40] Architecture with overlays highlighting the layers of the network affected by fine tuning.

*B.  Transfer Learning with Deep CNNs*

Transfer learning refers to extracting knowledge learnt from solving a *source* problem and applying it to an unrelated *target* task [41], [42]. In the particular context of pattern recognition and machine learning, this generally translates to making predictions for a data by reusing the models that have been trained for one or more unrelated tasks on a different dataset. Although the concept has been around since the mid 90's, its adoption has gained a lot of momentum recently with the revival of CNNs. The authors of [43] shed light on the fact that the features learnt by the bottom layers of deep CNNs trained on natural images are generic in the sense that they can be applied for an unrelated image task. Moreover, as we move from bottom to top of the architecture, the features gradually become more specific for the task at hand. This property of deep CNNs has been utilized for a variety of natural image analysis problems by rehashing models trained on large-scale datasets. Generally, the approach involves retaining the architecture of the original pre-trained CNN model and either using its first few layers as feature extractors, or incrementally updating the weights by resuming training using the data belonging to the task of interest. In this study, we apply the principles of transfer learning on the state-of-the-art network of InceptionV3 [40] that has been initially trained on the ImageNet dataset [14].

*1)   InceptionV3*
InceptionV3 is the third update to the original Inception network proposed in [25]. At the time, existing CNN architectures essentially consisted of multiple stacked convolution blocks (convolution + subsampling). Based on the premise that salient information in images can have a large variation in its scale and location, the authors of [25] proposed to replace traditional convolutional blocks with "Inception Modules". An inception module essentially performs max-pooling and convolutions at multiple scales (kernel sizes) and concatenates the feature maps from each of these operations. To reduce the large number of parameters, 1x1 convolutions are employed for dimensionality reduction. Furthermore, to tackle the issue of vanishing gradient associated with training deep networks, auxillary classifiers were used at intermediate points in the network, with the final loss function a weighted sum of all the different losses. Subsequent updates to the original architecture, named Inception V2 and V3 [40], aimed at further improving performance and computational efficiency by factorizing large convolutions, applying batch normalization on the auxillary classifiers and label smoothing. For our application, we sever the top 5 layers of the network and replace them with global average pooling layer, a fully connected layer with 1024 neurons and finally a softmax classifier which provides per-class probabilities. Following approaches were investigated for fine-tuning:

- Freezing all Inception convolution layers and training only the top layers. This is analogous to using the Inceptionv3 as an "off-the-shelf" feature extractor and using these features to train an MLP classifier. We refer to this approach as Inc-FE.

- Fine-tuning the last 1 or 2 inception blocks along with the top classification layers, while keeping the rest of the weights freezed. We refer to these two as Inc-1B and Inc-2B respectively in the paper.

Figure 3 highlights the layers of the architecture affected by the different approaches.

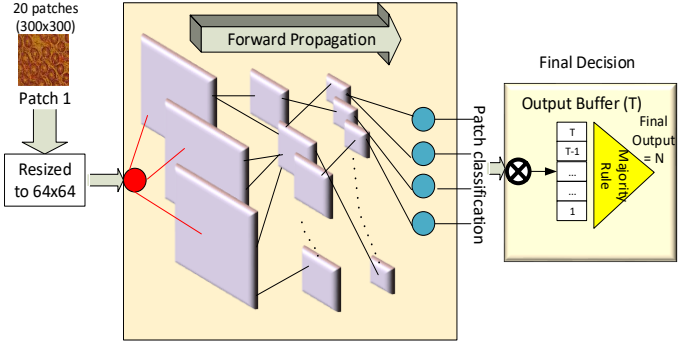

Figure 4 Proposed system architecture for Adaptive 2D CNNs

### C. Adaptive CNN Implementation

The proposed systematic approach for terrain segmentation is illustrated in Figure 4. As explained earlier, each of the 20 patches are used for training and classification. First each patch is downsized to 64x64 pixels to mimic low-resolution acquisition. Then its R,G,B channels are fed into the input layer of the CNN individually and the perform the final classification of the image, the majority rule is applied by averaging the individual class vectors of each patch.

In the proposed adaptive CNN implementation, there are two types of hidden layers: CNN layers into which conventional "convolutional" and "subsampling-pooling" layers are merged, and fully-connected (or MLP) layers. Neurons of the hidden CNN layers are, therefore, modified in such a way that each neuron is capable of both convolution and down-sampling (pooling). The intermediate outputs of each neuron are sub-sampled to obtain the final output of that particular neuron. The final output maps are then convolved with their individual kernels and further cumulated to form the input of the next layer neuron. In order to simplify the CNN analogy and to have the freedom of any input layer image dimension independent from the CNN parameters, neurons of the hidden "CNN layers" are modified as shown in Figure 5.

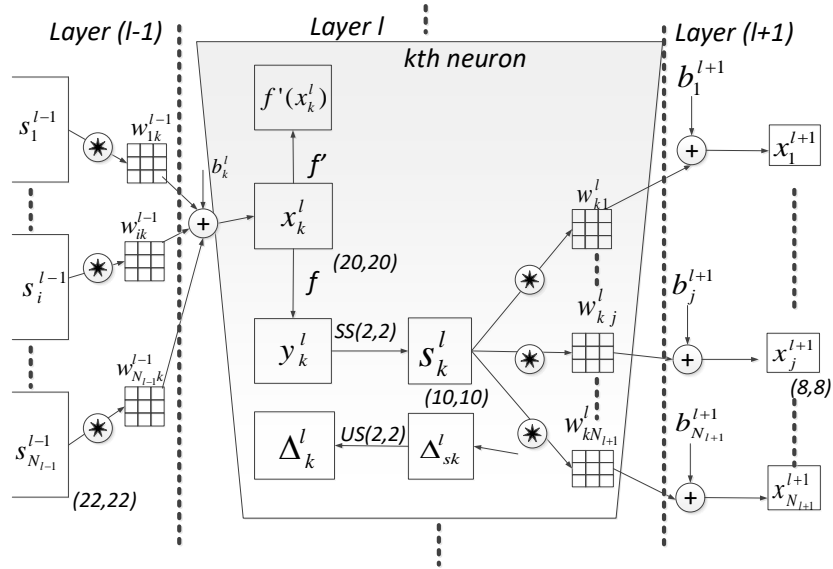

Figure 5 Adaptive CNN Implementation.

The final output of the $k^{th}$ neuron at layer $l$, $s_k^l$, is, therefore, the sub-sampled version of the intermediate output, $y_k^l$. The input map of the next layer neuron will be obtained by the cumulation of the final output maps of the previous layer neurons convolved with their individual kernels, as follows:

$$x_k^l = b_k^l + \sum_{i=1}^{N_{l-1}} conv2D(w_{ik}^{l-1}, s_i^{l-1}, 'NoZeroPad') \quad (1)$$

The number of hidden CNN layers can be set to any number. This ability is possible in this implementation because the sub-sampling factor of the output CNN layer (the hidden CNN layer just before the first MLP layer) is set to the dimensions of its input map, e.g., if the layer l+1 would be the output CNN layer, then the sub-sampling factors for that layer is automatically set to $ssx = ssy = 8$ since the input map dimension is 8x8. Besides the sub-sampling, note that the dimension of the input maps are gradually decreasing due to the convolution without zero padding. As a result of this, the dimension of the input maps of the current layer is reduced by ($K_x$-1, $K_y$-1) where $K_x$ and $K_y$ are the width and height of the kernel, respectively. The input layer is fed with the down-sampled patches each of which has three color channels (R,G,B). For a detailed mathematical breakdown of the backpropagation process, the reader is referred to the Appendix.

## V. EXPERIMENTS

### A. Setup

For transfer learning experiments, the deep CNN was programmed in Python using Keras library and the experiments were conducted using an NVIDIA K80 GPU with 12GB of memory on an Intel Xeon CPU E5-2690 processor running at 2.60 GHz with 64 GB of RAM. The implementation of the adaptive CNN was performed using C++ over MS Visual Studio 2015 in 64bit. This is a non-GPU implementation; however, Intel ® OpenMPI API is used to obtain multiprocessing with a shared memory. We used a computer with I7-4700MQ at 2.4GHz (8 CPUs) and 16GB memory for the training and testing purpose. The Adaptive CNN configuration parameters are presented in Table 1. For all experiments, we employed an early stopping training procedure with both setting the maximum number of BP iterations to 50 and the minimum training error to 8% to prevent over-fitting. For adaptive CNNs, we initially set the learning factor, $\varepsilon = 10^{-3}$ and then a global adaptation is applied during each BP iteration. This involves dynamically setting a proper learning factor for the iteration, i.e., if the train MSE decreases in the current iteration we slightly increase $\varepsilon$ by 5%; otherwise, we reduce it by 30%, for the next iteration.

Texture feature extraction algorithms were tested and evaluated in MATLAB. LibSVM toolbox [44] of MATLAB was utilized for SVM training and testing. The parameter estimation (cost and gamma) was done using grid search method.

Table 1
PARAMETERS FOR ADAPTIVE CNN

| Parameter Name | Value |
|---|---|
| Convolution filter size | 5 |
| Subsampling factor | 2 |
| No. of CNN layers | 3 |
| No. of MLP layers | 2 |

An 8-neighborhood schema was used for the LBP implementations. For the LPQ, a local window size of 3 was used with frequency estimation based on STFT with Gaussian window. A precomputed filter of window size 7 was applied to the rotation invariant LPQ descriptor. The statistics such as Contrast, Correlation, Energy, and homogeneity were computed from the GLCM matrix for the extraction of Haralick features.

### B. Performance Evaluation Metrics

We performed 5-fold cross validation over the dataset with the train-test split percentage set to 80%. Consequently, for each fold, 160 images and the corresponding 3200 patches were generated and used for training and the rest was used for testing. This allows us to evaluate the classifier over the entire dataset. Once the all patches of each image is classified by any classifier, the final decision is composed by the majority voting rule that is performed as follows: the values that represent the probabilities for belonging to each class obtained from the classifier are averaged over the patches and the final decision for that specific image is taken based on the highest probability obtained. Once the final decision is performed by a majority voting scheme over the patch classification results for each test image, the confusion matrices (CMs) are then cumulated to compute the final matrix, $CM_{Final}$.

Evaluation was performed separately for the two tasks of cancer detection and identification. For identification, we used the classification accuracy (Acc) computed over $CM_{Final}$ to quantify the overall performance. For the broader task of cancer detection, the $CM_{Final}$ was shrunk to a binary (2x2) matrix, $CM_{Binary}$, that has the numbers for normal and cancer categories. This was done by merging the classification results of the three cancer categories, HP, TA-LG and CA, into a single "cancer" class. Over $CM_{Binary}$ we computed the following standard metrics:

- *Accuracy* is the ratio of the number of correctly classified images to the total number of images, *Acc = (TP+TN)/(TP+TN+FP+FN)*;
- *Sensitivity (Recall)* is the rate of correctly classified cancer images among all cancer images, *Sen= TP/(TP+FN)*;
- *Specificity* is the rate of correctly classified normal images among all normal images, *Spe = TN/(TN+FP)*;
- *Positive Predictivity (Precision)* is the rate of correctly classified cancer images in all images classified as cancer, *Ppr= TP/(TP+FP)*.

Sensitivity and specificity are of particular importance when dealing with computer-aided diagnosis applications as they correspond directly to the percentage of correctly identified sick and healthy people, respectively.

Table 2
CLASSIFICATION ACCURACY (%) FOR TRADITIONAL METHODS WITH BINARY AND MULTICLASS CLASSIFICATION

| Method | Cancer Identification | | | Cancer Detection | | |
|---|---|---|---|---|---|---|
| | RBF | Linear | Poly. | RBF | Linear | Poly. |
| rLPQ | 67.5 | **74** | 71 | 84.5 | **87** | 83.5 |
| rLBP | 58 | 57.5 | **65.5** | 81 | 82.5 | **86** |
| Uniform rLBP | 56.5 | 54 | **59.5** | 80 | 79.5 | **83** |
| Haralick | 38 | 40 | **42.5** | 71 | 65 | **73** |
| rLPQ+ rLBP | 64 | 73.5 | **74** | 84 | 86 | **87** |

### C. Results and Comparative Evaluations

#### 1) Cancer Detection

For the traditional methods, Table 2 shows the detection accuracy for all combinations of textural features and kernels used for the classifier. Apart from the rLPQ, the polynomial kernel provides the best performance for all other texture features. Therefore, we keep the choice of kernel fixed to polynomial for the rest of this evaluation. Figure 6 provides a comprehensive comparison for all tested methods across multiple evaluation metrics. Among the traditional approaches, the rLPQ features and their combination with rLBP yielded the best overall performance whereas the Haralick features fair the worst. Furthermore, the transfer learning-based approaches provide quite similar results to each other, with an average classification accuracy of 90.68% and mean sensitivity of 94.6%. Inc-FE slightly edges out the other two approaches with respect to accuracy, specificity and precision, whereas Inc-1B provides the best sensitivity. Most critically, the CNN-based methods provide an average detection sensitivity of 94.5% which is about 5% more than the best performing traditional method.

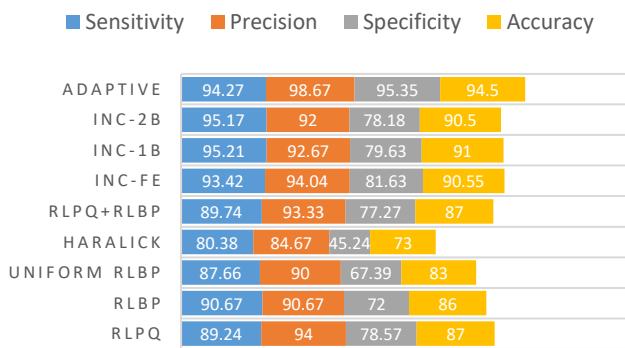

Figure 6 Cancer detection performance across multiple evaluation metrics.

Furthermore, among the CNN based methods, the proposed adaptive approach provides superior and most consistent detection results across all evaluation criteria and achieves a specificity level which is 16.81% higher than the next best performing CNN based method. It achieves a detection accuracy of 94.5% that is about 3.85% more than the top performing transfer learning-based method. In terms of sensitivity, Inc-1B with a percentage sensitivity of 95.21% yields the best results. However, it is only 1% higher than the levels achieved by the proposed adaptive approach.

*2) Cancer Identification*

As discussed earlier, cancer identification extends the detection problem to a multi-class classification where the cancer types are further categorized. The classification accuracies obtained by the traditional methods for each class are given in Table 2. Again, the polynomial kernel provided the best overall accuracy and was, therefore, used for all the comparative evaluations. Figure 7 illustrates the classification accuracies for the best traditional method, transfer learning and the proposed adaptive approach over each class. The accuracy for Class 2 (HP) is generally low across all methods, with an average of 61.6%. This can be attributed to the fact that although presence of hyperplastic polyps puts an individual at a higher risk of developing cancer [45], the polyps are mostly benign and are, at worst, precursors to colorectal tumors [46]. Furthermore, in our dataset, we observe that the inter-class similarity between Normal and HP class is visibly higher as compared with other classes. Figure 8 illustrates this visually. Moreover, the transfer learning-based methods; Inc-FE, Inc-1B and 1nc 2B, provide average identification accuracies of 83.5%, 87% and 84% respectively, each of which is superior to both the best traditional approach (74%) and the proposed adaptive approach (73.5%). Among the transfer learning methods, Inc-1B provided the best results, while the performance got worse for Inc-2B. This indicates that retraining more convolution blocks might lead to overfitting. As an overall trend, we can notice that the CNN-based methods, despite the low-resolution of input images (64x64 pixels), outperform the traditional methods, even when the latter use the full-resolution patch (300x300 pixels).

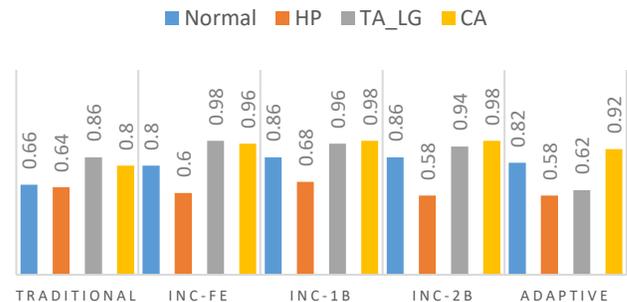

Figure 7 Cancer identification accuracies of the competing methods for each class.

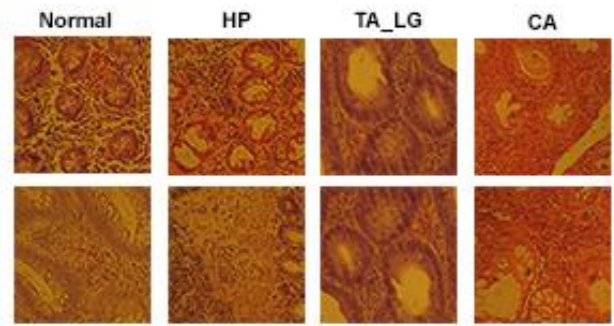

Figure 8 Sample patches from the QU-AHLI Dataset belonging to 4 classes.

VI. CONCLUSIONS AND FUTURE WORK

In this study, we investigate the performance of various machine-learning approaches for the task of colorectal cancer detection and identification especially when the data is scarce and the resolution is low. Our results reveal that CNN-based methods fair substantially better as compared to traditional machine learning approaches. Between the two CNN-based paradigms, deep vs. compact, we revealed that training an adaptive and compact CNN from scratch provides significantly better accuracy and consistent cancer detection results as compared to far deeper and complex fine-tuned models. For the more granular task of cancer identification, the proposed approach achieves classification results that are *at-par* with the traditional methods. However, the features learnt by deep models prove to be much more effective and provide a superior overall identification performance level (greater than 95%) for the three major cancer classes encountered. We have noted, however, that for the HP class, which is in fact a transition class between the normal and cancerous cells and thus exhibits a significant similarity to both counterparts, the identification performance degrades significantly for all approaches, as an expected outcome.

Our future work involves experimenting and comparing the proposed adaptive scheme on larger datasets, especially those compromising of multispectral images. We also aim to diversify our dataset by adding biopsy samples from even higher number of patients (e.g. > 500) and incorporating parameter-independent preprocessing measures.

## VII.    APPENDIX

### A. Back-Propagation for Adaptive CNNs

For an N-class problem, for each patch with its corresponding target and output vectors, $[t_1, \ldots, t_N]$ and $[y_1^L, \ldots, y_N^L]$, respectively, we are interested to find out the derivative of this error with respect to each individual network parameter (weights and biases). Let $l=1$ and $l=L$ be the input and output layers, respectively. The error (MSE) in the output (MLP) layer can be expressed as:

$$E = E(y_1^L, \ldots, y_{N_L}^L) = \sum_{i=1}^{N_L} (y_i^L - t_i)^2 \quad (2)$$

The BP formulation of the MLP layers is identical to the conventional BP for MLPs and hence skipped here. The BP training of the CNN layers is composed of 4 distinct operations as detailed below.

*1)    Inter-BP among CNN layers:* $\Delta s_k^l \overset{\Sigma}{\leftarrow} \Delta_l^{l+1}$

The basic rule of BP states: If the output of the k$^{th}$ neuron at layer *l* contribute a neuron *i* with weight $w_{ki}^l$ in the next layer *l*+1, the next layer neuron's delta $\Delta_i^{l+1}$ will contribute with the same weight to form $\Delta_k^l$ of the neuron in the previous layer *l*. This means:

$$\frac{\partial E}{\partial s_k^l} = \Delta s_k^l \overset{\Sigma}{\leftarrow} \Delta_i^{l+1}, \forall i \in \{1, N_{l+1}\} \quad (3)$$

where, *E* is the total error (MSE). Specifically:

$$\Delta s_k^l = \sum_{i=1}^{N_{l+1}} \frac{\partial E}{\partial x_i^{l+1}} \frac{\partial x_i^{l+1}}{\partial s_k^l} = \sum_{i=1}^{N_{l+1}} \Delta_i^{l+1} \frac{\partial x_i^{l+1}}{\partial s_k^l} \quad (4)$$

where

$$x_i^{l+1} = \cdots + s_k^l * w_{ki}^l + \cdots \quad (5)$$

It is obviously hard to compute the derivative directly from the convolution. Instead let us focus on a single pixel's contribution of the output $s_k^l(m,n)$, to the pixels of the $x_i^{l+1}(m,n)$ with the assumption of a 3x3 kernel for illustration simplicity.

$$x_i^{l+1}(m-1, n-1) = \cdots + s_k^l(m,n). w_{ki}^l(2,2) + \cdots$$

$$\boldsymbol{x_i^{l+1}(m-1, n) = \cdots + s_k^l(m,n). w_{ki}^l(2,1) + \cdots} \quad (6)$$

$$x_i^{l+1}(m+1, n+1) = \cdots + s_k^l(m,n). w_{ki}^l(0,0) + \cdots$$

This is illustrated on Figure 9 where the role of an output pixel, $s_k^l(m,n)$, over two pixels of the next layer's input neuron's pixels, $x_i^{l+1}(m-1, n-1)$ and $x_i^{l+1}(m+1, n+1)$ is indicated.

Considering the pixel as a MLP neuron that are connected to other MLP neurons in the next layer, according to the basic rule of BP one can then easily write the delta of $s_k^l(m,n)$ as follows:

$$\frac{\partial E}{\partial s_k^l}(m,n) = \Delta s_k^l(m,n)$$

$$= \sum_{i=1}^{N_{l+1}} \sum_{r=-1}^{1} \sum_{t=-1}^{1} \Delta_i^{l+1}(m+r, n+t). w_{ki}^l(1-r, 1-t) \quad (7)$$

If we generalize it for all pixels of $\Delta s_k^l$,

$$\Delta s_k^l = \sum_{i=1}^{N_{l+1}} conv2D(\Delta_i^{l+1}, rot180(w_{ki}^l), 'ZeroPad') \quad (8)$$

Note that this is a full convolution with zero padding by ($K_x$-1, $K_y$-1) zeros to each boundary of the $\Delta_i^{l+1}$ in order to achieve an equal dimensions (width and height) for $\Delta s_k^l$ and $\Delta_i^{l+1}$ with the $s_k^l$.

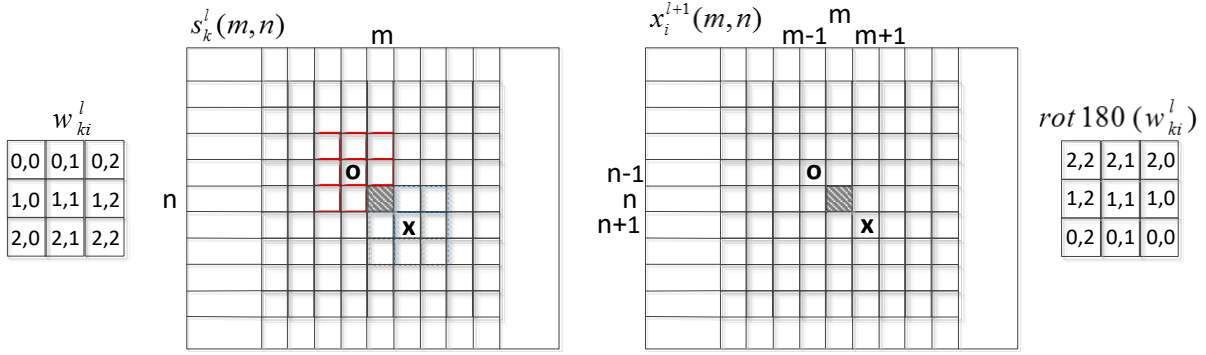

Figure 9: A single pixel's contribution of the output, $s_k^l(m,n)$, to the two pixels of the $x_i^{l+1}$ using a 3x3 kernel.

*2) Intra-BP within a CNN neuron:* $\Delta_k^l \leftarrow \Delta s_k^l$

Once the first BP is performed from the next layer, $l+1$, to the current layer, $l$, then we can further back-propagate it to the input delta. Let zero order up-sampled map be:

$$us_k^l = up_{ssx,ssy}(s_k^l),$$

then one can write:

$$\Delta_k^l = \frac{\partial E}{\partial x_k^l} = \frac{\partial E}{\partial y_k^l}\frac{\partial y_k^l}{\partial x_k^l} = \frac{\partial E}{\partial us_k^l}\frac{\partial us_k^l}{\partial y_k^l} f'(x_k^l) \quad (9)$$

$$= up(\Delta s_k^l)\beta\, f'(x_k^l)$$

where $\beta = (ssx, ssy)^{-1}$ since each pixel of $s_k^l$ was obtained by averaging $ssx.ssy$ number of pixels of the intermediate output $y_k^l$. If "maximum pooling" is used instead of an "average pooling", then Eq. (9) should be adapted accordingly.

*3) BP from the first MLP layer to the last CNN layer*

As illustrated in Figure 10, the last (or output) CNN layer is connected to the 1st MLP layer and hence the outputs of this layer's CNN neurons, $s_k^l$, are scalars. In other words, $s_k^l$ and of course, $\Delta s_k^l$ are now all scalars and to achieve this recall that the subsampling factors, ssx and ssy, in this particular layer are all set to the dimensions of the input map (ssx= 8, ssy= 8) as in the figure. Similarly the weights of this CNN layer neurons, $w_{ki}^l$, are also all scalar and instead of convolution, scalar multiplication is performed as in a regular MLP. So from MLP layer to the CNN layer, the regular (scalar) BP is simply performed as in Eq. (10).

$$\frac{\partial E}{\partial s_k^l} = \Delta s_k^l = \sum_{i=1}^{N_{l+1}} \frac{\partial E}{\partial x_i^{l+1}}\frac{\partial x_i^{l+1}}{\partial s_k^l} = \sum_{i=1}^{N_{l+1}} \Delta_i^{l+1} w_{ki}^l \quad (10)$$

and intra BP to get: $\Delta_k^l \xleftarrow{BP} \Delta s_k^l$ is identical to a regular BP for MLPs.

$$\Delta_k^l = \frac{\partial E}{\partial x_k^l} = up(\Delta s_k^l)\,\beta\, f'(x_k^l) \quad (11)$$

Finally, the weight and bias sensitivities, too, are identical to a regular MLP's:

$$\frac{\partial E}{\partial w_{kj}^l} = \frac{\partial E}{\partial x_j^{l+1}}\frac{\partial x_j^{l+1}}{\partial w_{kj}^l} = \Delta_j^{l+1} s_k^l \quad (12)$$

$$\frac{\partial E}{\partial b_k^{l+1}} = \frac{\partial E}{\partial x_k^{l+1}}\frac{\partial x_k^{l+1}}{\partial b_k^{l+1}} = \Delta_k^{l+1}$$

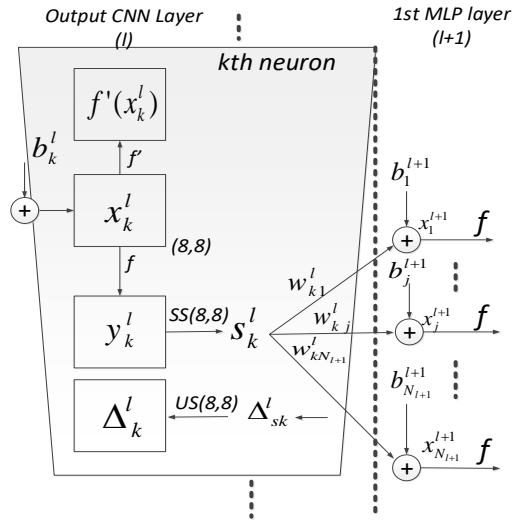

Figure 10: The output CNN layer and the 1st MLP layer.

*4) Computation of the Weight (Kernel) and Bias Sensitivities*

As in the regular BP on MLPs, the delta of the $i^{th}$ neuron at layer $l+1$, $\Delta_i^{l+1}$ will be used to update the bias of that neuron and all weights of the neurons in the previous layer connected to that neuron.

$$x_i^{l+1} = b_i^{l+1} + \cdots + y_k^l w_{ki}^l + \cdots$$

$$\therefore \frac{\partial E}{\partial w_{ki}^l} = y_k^l \Delta_i^{l+1} \text{ and } \frac{\partial E}{\partial b_i^{l+1}} = \Delta_i^{l+1} \quad (13)$$

The update rule of the conventional BP states: The sensitivity of the weight connecting the $k^{th}$ neuron in the current layer to the $i^{th}$ neuron in the next layer depends on the output of the current layer neuron, and the delta of the next layer neuron. For hidden neurons in a CNN layer we need to follow a similar approach to find out weight and bias sensitivities.

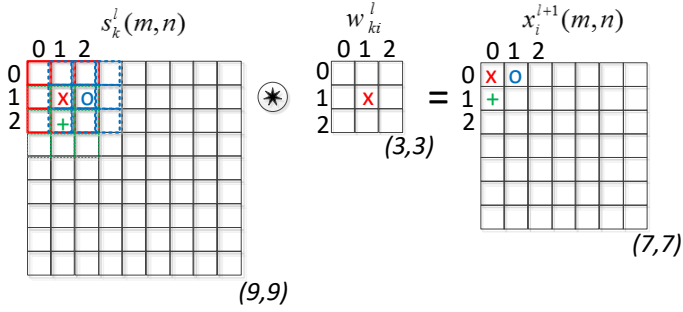

Figure 11: Convolution of the output of the current layer neuron, $s_k^l$, and kernel, $w_{ki}^l$, to form the input of the $i^{th}$ neuron, $x_i^{l+1}$, at the next layer, $l+1$

Figure 11 illustrates the convolution of the output of the current layer neuron, $s_k^l$, and kernel, $w_{ki}^l$, to form the input of the $i^{th}$ neuron, $x_i^{l+1}$, at the next layer $l+1$. So if we focus on the contribution of each kernel element over the output, in analytical form one can write:

$$x_i^{l+1}(0,0) = .. + w_{ki}^l(0,0)s_k^l(0,0) + w_{ki}^l(0,1)s_k^l(0,1)$$
$$+ w_{ki}^l(1,0)s_k^l(1,0) + ..$$

$$x_i^{l+1}(0,1) = .. + w_{ki}^l(0,0)s_k^l(0,1) + w_{ki}^l(0,1)s_k^l(0,2)$$
$$+ w_{ki}^l(1,0)s_k^l(1,1) + ..$$

$$\mathbf{x_i^{l+1}(1,0) = .. + w_{ki}^l(0,0)s_k^l(1,0) +}$$
$$\mathbf{w_{ki}^l(0,1)s_k^l(1,1) + w_{ki}^l(1,0)s_k^l(2,0) + ..} \quad (14)$$

….

$$x_i^{l+1}(m,n) = .. + w_{ki}^l(0,0)s_k^l(m,n) + w_{ki}^l(0,1)s_k^l(0,n+1)$$
$$+ w_{ki}^l(1,0)s_k^l(m+1,n) + ..$$

$$x_i^{l+1}(m,n) = \sum_{r=-1}^{1}\sum_{t=-1}^{1} w_{ki}^l(r+1,t+1)s_k^l(m+r,n+t) + ..$$

Since each weight (kernel) element is used in common to form each neuron input $x_i^{l+1}(m,n)$, the derivative will be the cumulation of delta- output product for all pixels. i.e.,

$$\frac{\partial E}{\partial w_{ki}^l(r,t)} = \sum_m \sum_n \Delta_i^{l+1}(m,n)s_k^l(m+r,n+t) \quad (15)$$

$$\Rightarrow \frac{\partial E}{\partial w_{ki}^l} = conv2D(s_k^l, \Delta_i^{l+1}, 'NoZeroPad')$$

Similarly, the bias for this neuron, $b_k^l$, contributes to all pixels in the image (same bias shared among all pixels), so its sensitivity will be the cumulation of individual pixel sensitivities as expressed in Eq. (16).

$$\frac{\partial E}{\partial b_k^l} = \sum_m \sum_n \frac{\partial E}{\partial x_k^l(m,n)} \frac{\partial x_k^l(m,n)}{\partial b_k^l} =$$
$$\sum_m \sum_n \Delta_k^l(m,n) \quad (16)$$

As a result, the iterative flow of the BP for each patch in the training set can be stated as follows:

**Algorithm 1. Iterative Flow of BP for Adaptive CNN**

Initialize weights (kernels) and biases (e.g., randomly, U(-0.1, 0.1)) of the CNN.

**FOR** each BP iteration (*t=1:iterNo*) **DO**:

  **FOR** each patch, *p*, in the train set, **DO**:

   **FP**: Forward propagate from the input layer to the output layer to find output of each neuron at each layer,. $y_i^l, \forall i \in [1, N_l]$ and $\forall l \in [1,L]$

   **BP**: Compute delta error at the output (MLP) layer and back-propagate it to first hidden CNN layer to compute the delta errors, $.\Delta_k^l, \forall k \in [1, N_l]$ and $\forall l \in [2, L-1]$

   **PP**: Post-process to compute the weight and bias sensitivities using Eqs. (15) and (16).

   **Update**: Update the weights and biases with the (cumulation of) sensitivities found in (c) scaled with the learning factor, $\varepsilon$, as follows:

$$w_{ik}^{l-1}(t+1) = w_{ik}^{l-1}(t) - \varepsilon \frac{\partial E}{\partial w_{ik}^{l-1}} \quad (17)$$

$$b_k^l(t+1) = b_k^l(t) - \varepsilon \frac{\partial E}{\partial b_k^l}$$